\documentclass[letterpaper, 10 pt, conference]{ieeeconf}  

\IEEEoverridecommandlockouts                              

\usepackage{graphicx} 
\usepackage{amsmath} 
\usepackage{amssymb}  
\usepackage{hyperref}
\usepackage{algorithm}
\usepackage{algpseudocode}
\usepackage{cite}
\usepackage{tikz}
\usepackage{siunitx}

\newcommand\submittedtext{%
  \footnotesize This work has been submitted to the IEEE for possible publication. 
  Copyright may be transferred without notice, after which this version may no longer be accessible.}

\newcommand\submittednotice{%
  \begin{tikzpicture}[remember picture,overlay]
    \node[anchor=south,yshift=15pt] at (current page.south) 
    {\parbox{\dimexpr0.8\textwidth}{\centering\submittedtext}};
  \end{tikzpicture}
}

\title{\LARGE \bf
Idiobionics: The Unification\\of Privacy and Intelligent Robotic Prostheses
}

\author{Kwesi Afari Darfoor,$^{1,2}$ Patrick M. Pilarski,$^{1,2,3}$  and Bailey Kacsmar$^{1,2}$ 
\thanks{*This work was supported in part by the Alberta Machine Intelligence Institute (Amii), the Canada CIFAR AI Chairs Program, the Canadian Artificial Intelligence Safety Institute (CAISI), the National Science and Engineering Research Council (NSERC), Alberta Innovates, and the Government of Alberta. $^{1}$Department of Computing Science,
        University of Alberta, Edmonton, Canada; $^{2}$Alberta Machine Intelligence Institute (Amii), Edmonton,
        Canada; $^{3}$Department of Medicine, University of Alberta, Edmonton, Canada.}%
\thanks{
        {\tt\small \{darfoor,pilarski,kacsmar\}@ualberta.ca}}
}
\begin{document}

\maketitle
\submittednotice
\thispagestyle{empty}
\pagestyle{empty}

\begin{abstract}

The human body is at the center of a growing family of technologies designed to tightly and persistently couple biological and digital systems. Robotic prostheses are a representative example of this tight coupling. Also referred to as bionic limbs, robotic prostheses are devices that support people who have lost limbs in pursuing daily life activities such as walking and grasping objects.
Bionic limbs  are now perceptive and responsive owing to their integration with advanced sensors and artificial intelligence-based control approaches. Consequently, 
such robotic prostheses can now be viewed as semiautonomous wearable robotic systems that can co-adapt with their users. 
However, the same sensing and control advancements that increase the capability of robotic prostheses also introduce threat vectors that could be exploited by malicious entities to violate the privacy of users. To fully realize the benefits of next-generation bionic limbs, we maintain it is important to directly understand and address these privacy risks and the barriers they might present to user adoption. 
This paper therefore introduces a new line of inquiry we term {\em idiobionics} to holistically investigate issues at the intersection of privacy and intelligent bionic limbs. As the main contribution of this paper, we define idiobionics, ground it in related literature, and provide preliminary evidence showing and discussing potential adversarial attacks that could exploit intelligent bionic limb designs. We then contribute a curated list of open research questions within idiobionics that are relevant to researchers in wearable robotics and other human-facing autonomous systems. 
We expect that idiobionics research will help unlock the full potential of robotic prostheses and related bionic devices.

\end{abstract}

\section{INTRODUCTION}

We are currently in an unprecedented technological era where devices are integrated with the human body, transforming a user into a human-machine entity in what is called the Internet of Bodies (IoB) ~\cite{Lee2020, Rani, neal}.  First introduced in 2014, the IoB encompasses a wide range of devices, from advanced internal devices such as pacemakers and cochlear implants, to popular external wearables such as fitness trackers~\cite{Lee2020}. Driven by increasing consumer adoption, these devices have become an integral component of daily life, evolving into tools that users rely on to function and perform various activities~\cite{Lee2020, Rani}. In this paper, we focus on a specific and representative example of an IoB device: robotic prostheses, also termed bionic limbs, intended for long-term daily use by people with amputations or other forms of limb difference. 

Bionic limbs are integral to the lives of many individuals who have lost a limb due to congenital abnormalities or traumatic injuries like motorcycle and workplace accidents~\cite{ mcdonald2021global}. These robotic devices are engineered to allow such individuals to regain functional movement tailored to their specific amputation level~\cite{frossard2021future} and therefore carry out various Activities of Daily Living (ADLs) including walking and grasping objects~\cite{law1989critical}. By integrating biomechanical principles and robotic technology, bionic limbs aim to improve mobility, independence, and overall quality of life~\cite{frossard2021future}, making them a vital solution for the millions of individuals with both upper and lower limb difference worldwide~\cite{websiteamputeecoalition, mcdonald2021global}.

Current bionic limbs, both commercially available and research prototypes, are designed to be increasingly autonomous and adaptive devices capable of intelligently interpreting the individual's movement intent and executing precise commands~\cite{Campbell2025, fleming2021myoelectric, kim2023machine, williams2022composite}. The adaptability of bionic limbs relies on a system with two integrated components: a control component and a perception component. The perception component uses various sensors to acquire data about the user and the environment. This primarily includes the Electromyographic (EMG) sensor which measures electrical activity from muscles alongside accelerometers and gyroscopes that capture linear and rotational movement data respectively~\cite{lamsellak2023hand, Kurniawan, osborn2020sensing, Su, Yadav}. These sensors provide comprehensive data that are fed into the control component now powered by machine learning algorithms such as Support Vector Machines (SVM), random forests, Linear Discriminant Analysis (LDA), and neural networks~\cite{fleming2021myoelectric, williams2022composite, kim2023machine}. These models translate the data into actionable commands for the bionic limb~\cite{fleming2021myoelectric, williams2022composite, kim2023machine}. Active research on continual learning, incremental learning, and reinforcement learning strategies ~\cite{Campbell2025, wu2021using, nowak2023simultaneous, egle2023preliminary} is also being conducted to help eliminate the need for frequent user calibration and to accommodate for changes over time, such as long-term habituation and transient states like fatigue~\cite{lamsellak2023hand, cimolato2020, Su, websitecoapt}. This advanced sensing and computing  functionality is expected to ultimately allow bionic limbs to adapt to a user’s unique movement patterns in real time, facilitating an intelligent system where the human and their device jointly learn and improve so as to achieve the goal of functional dexterity~\cite{Campbell2025,Dawson2024}.

Although these technological advancements promise functional benefits to users, they also introduce vulnerabilities that might compromise user privacy. This, in turn, could undermine adoption and hinder users from fully experiencing the functional benefits of modern bionic limbs~\cite{kuberkar2020factors, Heerink2010, choudhury2023investigating}. In anticipation of these privacy and security related considerations, we  introduce what we believe to be a necessary and emergent line of inquiry into how integrated technological features in bionic limbs can enable adaptability while not inadvertently causing harm to users. We term this pursuit \emph{idiobionics}. As a first contribution of this manuscript, we formally define idiobionics as a new interdisciplinary scope of research that is relevant to the community of wearable autonomous robotics researchers (Section~\ref{sec: idiobionics}). We then present preliminary evidence of inherent privacy risks within an upper-limb bionic device as a representative example (Section~\ref{sec: empirical evidence}). In Section~\ref{sec: privacy outlook}, we discuss the implications of our findings along with potential risks for bionic limb technology. Finally in Section~\ref{sec: future direction} of this paper, we offer plausible future research directions within idiobionics research.

\section{Idiobionics}
\label{sec: idiobionics}

Establishing consumer trust is a first step to realizing the benefits of autonomous and adaptive technologies. 
Lack of user trust is known to hinder the widespread adoption of adaptive technology, ultimately rendering advancements and innovations in the field futile since the intended users are not going to be actively benefiting from them~\cite{kuberkar2020factors, Heerink2010, choudhury2023investigating}. Users are less likely to trust and thus adopt adaptive devices if using the device necessitates surrendering their valued privacy rights~\cite{lenhart2023you, kobayashi2016ethical}. 

It is our assertion that the same technological integrations that allow for adaptability in bionic limbs also introduce vulnerabilities and threat vectors that could be exploited by adversaries to deduce sensitive information about users, such as their daily activities~\cite{orlosky2019look, Velykoivanenko, Salehzadeh, weiss2019smartphone, shahmohammadi2017smartwatch}, specific demographic and anthropometric attributes like age and height~\cite{Kroger, riaz2015one, hoffmann2018estimating, yanai2016estimating}, or health status~\cite{lippi2020estimation, dubbert2002obesity}, thereby violating privacy. Such unauthorized information deduction could lead to serious privacy risks. These include discriminatory practices like being denied or charged higher insurance premiums based on inferred health risks derived from their activity data~\cite{websitetheatlantic}. Additional privacy risks include the potential for targeted physical crimes such as robberies and burglaries facilitated by determining a user's daily routines~\cite{logan2017stalking, websitenyt}.

Concerns regarding the privacy implications of devices tightly coupled with the human body such as bionic limbs are not a recent development. Meghan Neal first introduced the term \emph{`Internet of Bodies'} in 2014; her article highlighted the privacy issues that emerge as a result of increasingly direct human-machine integration~\cite{neal}. A decade later, other works have gone on to further this discussion on the privacy issues within the IoB landscape~\cite{orlosky2019look, Salehzadeh, shahmohammadi2017smartwatch, Velykoivanenko, Weiss}, including substantive legal considerations~\cite{Matwyshyn, leenes2018data}.

However, while prior research addresses the privacy risks of popular IoB devices like fitness trackers and smartwatches, there is a critical gap regarding bionic limbs~\cite{orlosky2019look, Salehzadeh, shahmohammadi2017smartwatch, Velykoivanenko, Weiss}. Bionic limbs which are classified as medical devices are persistently and in some cases invasively affixed to the human body. This means that unlike many personal electronic devices, the clinical, financial, and regulatory considerations around prescribed prosthetic technologies make bionic limbs challenging or impossible for a user to replace. A smartwatch with a known hardware privacy vulnerability might be something a user could replace; an insurance-provided \$100k prosthetic limb, customized to an individual and with limited commercial alternatives, is often not replaceable. Therefore, broader and more comprehensive insight is required into the privacy risks stemming from the deep integration of adaptive bionic limbs with human users. This proactive measure will thus support a privacy-by-design approach of future sophisticated prostheses where privacy protections are incorporated into designs from the onset.


To address this deficiency in comprehensive privacy insight, we now describe a shared line of inquiry we call idiobionics, defined as interdisciplinary research dedicated to enhancing privacy in autonomous, semiautonomous, or adaptive bionic limbs. We see idiobionics research as 
a means to accelerate progress on intelligent bionic devices, while also facilitating the commercialization of innovative technology by enabling us to preemptively address privacy concerns. By prioritizing idiobionics research, we can ensure that users can trust and adopt adaptive bionic limbs without fear of privacy violations, especially given the tight coupling of robotic prostheses with the human body. 

Idiobionics research will ensure that companies can build compliance into the core architecture of intelligent bionic limbs, 
streamlining regulatory approval and accelerating the path to commercial viability. Ultimately, we expect idiobionics will involve technical assessments to identify vulnerabilities and threat vectors within autonomous and adaptive bionic limbs that could be exploited for privacy violations. Thus, such evaluations will enable preemptive mitigation of privacy risks. Furthermore, we believe this new line of inquiry will by necessity integrate extensive qualitative and quantitative research that engages with all relevant stakeholders, including users.

By example, and as a first step within idiobionics research, we now present preliminary experimentation into a specific application of upper-limb bionic devices for individuals with transradial amputations (amputations below the elbow). This work serves as an initial step toward further investigation of privacy concerns, both within the bionic limb domain and across the wider landscape of bionic devices.

\section{Empirical Evidence}
\label{sec: empirical evidence}
As a representative example, in this section we show that the risk of an adversary inferring a user's activity from their accelerometer data through what we call an Activity Inference Attack (AIA), is feasible in a setting congruent with transradial bionic limb use, i.e., use of bionic limbs for amputations occurring below the elbow~\cite{darfoor}. This empirical evidence not only demonstrates the feasibility of privacy attacks in this domain but also provides technical insight into potential safeguards against them. In this section, we describe our experimental setup, including our data collection process and the analytical methods employed, which consists of both supervised and unsupervised machine learning approaches.

We collected forearm movement data using a tri-axial accelerometer, which measures movement along three axes in 3D space. This accelerometer was situated within a clinical-grade electromyography sensor. Accelerometer data was recorded in accordance with ethical guidelines approved by the institution's research ethics board. This study included twelve participants aged eighteen to sixty who self-selected to participate, provided written informed consent, and reported no physical limitations or pre-existing medical conditions that would prevent them from safely performing the required physical activities. Each participant was fitted with a tri-axial accelerometer above the wrist flexor muscle belly, which is located on the inner forearm, just below the elbow. The accelerometer captured movement data at 148\unit{\hertz} (148 data points recorded every second) as participants performed four controlled activities: walking, jogging in place, sitting, and standing. Each activity was performed one after the other for thirty seconds each within a single session in a lab environment and separate data were recorded for each. 

Once this data collection process was complete, the resulting dataset was used to assess the feasibility of conducting an activity inference attack. We explored both a supervised and an unsupervised learning approach. In our supervised learning approach, we employed a transfer learning strategy~\cite{torrey2010transfer}, using the HarNet10 model~\cite{yuan2024}, which is a publicly available pre-trained model. With twelve participants, the model was finetuned twelve different times. Each time the model was finetuned, data from one participant was left out for attacking, while data from the remaining eleven participants were used for finetuning, resulting in twelve different models. Each participant served as a training case or target for our models. After each of the twelve experimental runs, the model produced a prediction accuracy score. This resulted in twelve accuracy scores across the twelve runs. The average of all twelve accuracies was then computed to provide a robust overall performance metric for attacking the different targets.

\begin{figure}[t!]
  \vspace{2mm}
  \centering
  \includegraphics[width=0.80\linewidth]{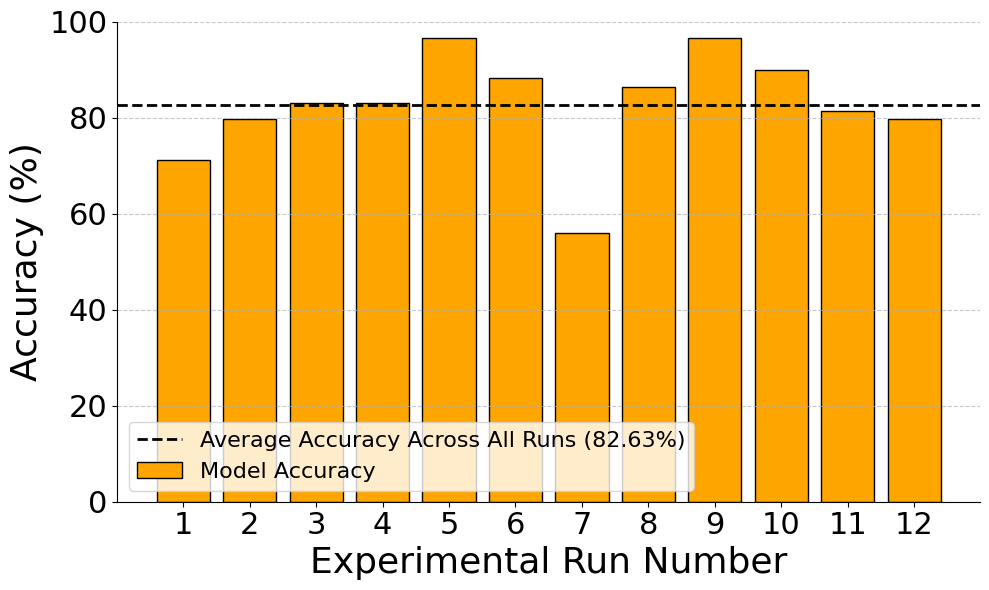}
  \caption{We illustrate the feasibility of using a publicly available pre-trained model for an Activity Inference Attack on bionic limb users with  an average attack accuracy of 83\% across twelve experimental runs.}
  \label{fig:results_overall}
\end{figure}

In Figure~\ref{fig:results_overall}, we observe that the average of the attack accuracy across all twelve experimental runs was 83\% when predicting which activity the user was performing. This accuracy level is high enough to conclude that an adversary may leverage a publicly available model previously pre-trained with accelerometer data to launch a successful activity inference attack. Figure~\ref{fig:results_overall} also further highlights substantial variability in model accuracy across participants. While the model distinguishes activities with high accuracy for some individuals (e.g., participants with ID5 and ID9 at 96\%), it performs significantly worse for others (e.g., participant with ID7 at 57\%). This variation suggests the presence of a subgroup of participants whose activity patterns are more easily identifiable, making them more susceptible to activity inference attacks. Therefore, designing protections with this subgroup in mind may yield protections that generalize to users with less distinctive activity patterns.

For our unsupervised learning approach, we used clustering algorithms to determine the extent to which different activities captured by the forearm-mounted accelerometer cluster in a feature space. To achieve this we first computed the magnitude of the recorded accelerometer data~\cite{lee2017human, al2023smartphone} (see lines three and four of Algorithm~\ref{alg:feature_extraction}). Then, using a non-overlapping sliding window technique to segment the data into two-second windows, we extracted relevant features from each resulting window~\cite{al2023smartphone, krishnan2014activity, banos2014window}. These features included statistical features such as mean, standard deviation, and kurtosis~\cite{sudhakar2021actid}, which were extracted from the time domain, as well as features such as dominant frequencies and spectral centroid~\cite{Dargie} which were extracted after transforming the data into the frequency domain. The extracted features were subsequently standardized~\cite{milligan1988study} and reduced to two dimensions using Principal Component Analysis (PCA)~\cite{jolliffe2011principal}. The resulting feature vectors were then fed into five different clustering algorithms to assess their performance in grouping similar data points.

\begin{algorithm}[b!]
\caption{Feature Extraction from Accelerometer Data}
\label{alg:feature_extraction}
\begin{algorithmic}[1]
\Statex Sampling rate $f_s=148$\unit{\hertz}, window duration $T=2$\unit{\second},  $N=3$

\State $W \gets f_s \cdot T$ \Comment{Window size in samples}
\State Load accelerometer data $D = \{X,Y,Z\}$

\For{each sample in $D$}
    \State Compute magnitude $M \gets \sqrt{X^2 + Y^2 + Z^2}$
\EndFor

\For{each activity}
    \State Segment $M$ to non-overlapping windows of size $W$
    \For{each window}
        \State Compute time-domain features \Comment{mean, std, skewness, kurtosis, min, max, median}
        \State Estimate power spectrum using Welch’s method
        \State Extract top $N$ frequency peaks
        \State Compute spectral centroid and band powers
        \State Concatenate all features into one feature vector 
        \State Append to feature list $X$
    \EndFor
\EndFor

\State Standardize $X$
\State Apply Principal Component Analysis to reduce $X$ to two dimensions ($X_{\text{PCA}}$)
\State \Return Feature Matrix $X_{\text{PCA}}$ \Comment{The $PC_1$ and $PC_2$ scores}

\end{algorithmic}
\end{algorithm}

The five algorithms used in the experiment were K-Means, DBSCAN (Density-Based Spatial Clustering of Applications with Noise), Gaussian Mixture Models (GMMs), OPTICS (Ordering Points To Identify the Clustering Structure), and Agglomerative Clustering. Out of these five algorithms, Gaussian Mixture Models (GMMs) was the best performing algorithm in the attack~\cite{darfoor}. To interpret the cluster results, an adversary can utilize a publicly available accelerometer dataset recorded from a similar location to see how it naturally clusters by activity. Alternatively, a more powerful adversary could use a high fidelity system which is costly, to collect and cluster their own validation data.

\begin{figure}[t!]
  \vspace{2mm}
  \centering
  \includegraphics[width=0.70\linewidth]{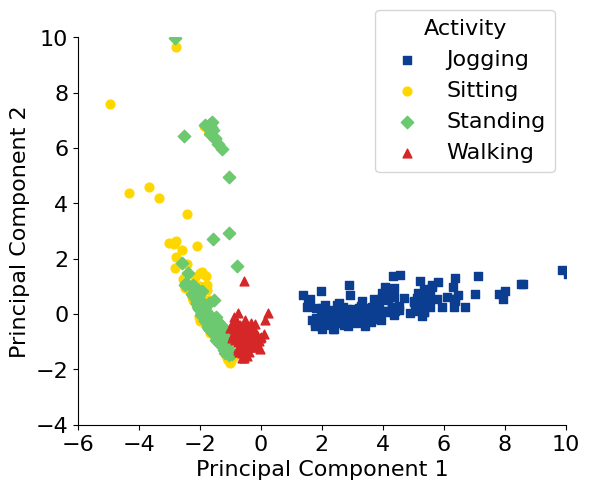}
  \caption{Distribution of the ground truth data captured from the forearm using a tri-axial accelerometer. Jogging shows clear separability, whereas walking data partially overlaps with the sitting and standing activity clusters.}
  \label{fig:trigno_true}
\end{figure}

\begin{figure}[t!]
  \centering
  \includegraphics[width=0.70\linewidth]{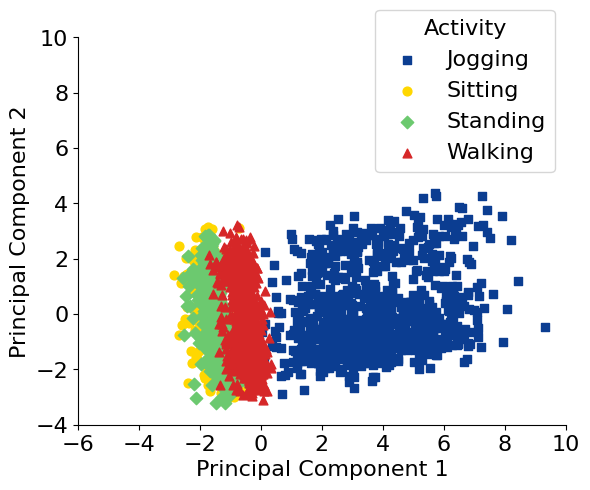}
  \caption{An adversary could utilize forearm accelerometer data from a smartphone which form clusters similar to those from the tri-axial accelerometer. Jogging consistently forms the most distinct cluster, while walking exhibits slight overlap with sitting and standing activities.}
  \label{fig:smartphone_true}
\end{figure}

However, in the absence of those two strategies, we demonstrate that an adversary could use easily available consumer-grade devices such as a smartphone to interpret the results. By placing a smartphone at a similar location as the tri-axial accelerometer which is around the wrist flexor region (generally right below the inner elbow), we collected accelerometer data while participants performed the four activities. We observe that the resulting smartphone data clusters in a similar pattern as the data collected by the tri-axial accelerometer as shown in Figure~\ref{fig:trigno_true} and Figure~\ref{fig:smartphone_true}.

\section{Privacy Risks Outlook}
\label{sec: privacy outlook}

Our findings in Section~\ref{sec: empirical evidence} demonstrate just one of many ways adversaries could use data from a common sensor to determine a user's daily routine and habits. We expect similar vulnerabilities to present themselves in a way that scales with the number and type of sensors in a given combination of bionic limb components and their supporting technology. 

A natural first privacy risk for consideration, that
unrestricted access to bionic limb sensor streams might expose, relates to user activity. As noted above, adverse outcomes related to the presence, absence, or timing of activity could include crimes such as targeted burglaries and theft~\cite{websitenyt}.
Activity data from a bionic limb could also be used to deduce a user's vulnerability to health issues, such as cardiovascular diseases based on the correlation between physical activity patterns and health risk~\cite{dubbert2002obesity, lippi2020estimation,langhammer2018importance, li2012physical}. This affords bad actors the potential to exploit or sell user health information, e.g., to third party entities such as pharmaceutical companies for profit~\cite{websitecbcnews, websitecbcnews2024, spithoff2025primary, websitetheguardian}.

Sensitive demographics (such as user age) are a second type of personal information we anticipate that the sensors normally embedded within bionic limbs could be exploited to deduce. Again using accelerometer data as an example, sensor streams related to limb movement can reveal a person's age by making visible changes in gait patterns (changes in the way a person walks) and variations in physical activity intensity or frequency over an extended period~\cite{pyrkov2018extracting, rahman2019deep, hoffmann2018estimating, riaz2015one}. Since gait analysis is primarily influenced by lower-body movement~\cite{tao2012gait}, we expect the feasibility of inferring age through changes in gait patterns from accelerometer data to pose a greater privacy risk to users of lower-limb bionic prostheses than to those using upper-limb devices.

Perhaps surprisingly, a third type of personal information that might be sampled by an adversary is the speech produced by a bionic limb user. Prior research has shown that sensors such as accelerometers can be exploited in speech-related privacy attacks, exposing spoken words, speaker gender, and the identity of specific speakers when media are played from a smartphone's loudspeaker~\cite{Anand, de2023improved, Weigao, michalevsky2014gyrophone, OwuEmma}. 
Such an attack is facilitated by the accelerometer's sensitivity to subtle vibrations. Due to the tight coupling of a bionic limb with its human user, we expect speech-targeting attacks exploiting embedded vibration-sensitive sensors or actuators to pose a transferable risk. We also highlight this as a lesson to carefully consider the integration of sound features in bionic limbs to avoid introducing new vulnerabilities that could be exploited in privacy attacks.

Other risks have been thoroughly documented in the smartphone and smartwatch domain, which are a part of the IoB landscape, showing that device movement can also be exploited to infer a user's location~\cite{Owusu, Kroger}, keystrokes including PINs and passwords~\cite{OwuEmma, Kroger}, and height~\cite{Kroger, Weiss, yanai2016estimating, riaz2015one}. Since smartwatches share a similar placement with bionic limbs in cases of wrist disarticulation~\cite{chow2023transradial}, there is an opportunity to determine whether the privacy risks documented in the smartwatch domain extend to bionic limbs.


While we have placed our focus on one common sensor, the accelerometer is not the only sensor embedded in an upper or lower bionic limb to pose potential privacy risks to users. Of note, any gyroscope and EMG sensors within a device are also capable of exposing sensitive information about users. Research has demonstrated that, as with accelerometer data, gyroscope data also pose powerful identity-based privacy risks since they could be used for identification and authentication purposes via gait analysis~\cite{OwuEmma}. Furthermore, the ability of EMG sensors to record muscle activity poses an unexpected but potentially direct privacy risk, providing what might be a variant of keystroke attacks~\cite{Ruide2017, arteaga2020emg, georgi2015recognizing}. Research shows that EMG data from   a commercially available non-medical myoelectric wearable, particularly when combined with the same device's accelerometer data, can facilitate the unauthorized inference of passwords and PINs~\cite{Ruide2017}. This may be relevant for future users of more dexterous forearm prostheses, e.g., if they have residual forearm muscles that activate as they engage with systems like a keypad or touch screen while they move their bionic limb to enter a passcode.

We also note here that privacy considerations transfer to related rehabilitation and assistive technology domains. One example closely related to bionic limbs are exoskeletons, which differ in that they  support existing, intact limbs rather than replacing them entirely~\cite{du2021review, chen2013review}. These devices may support upper limbs, lower limbs, or an individual's torso and hips. Exoskeleton use cases have already expanded beyond medical fields and into recreational and industrial applications. Thus, their associated privacy risks are expected to have a much broader future reach within society. Commercial exoskeletons such as Hypershell X by Hypershell Tech~\cite{websitehypershell} and the MO/Go by Arc'teryx and Skip~\cite{websitearcteryx} promise enhanced performance during outdoor activities such as mountain climbing. Recreational exoskeletons, as well as those utilized in medical settings to help spinal cord injuries or stroke users regain motor function, are embedded with sensors such as accelerometers, gyroscopes and pressure sensors, which introduce many of the above-noted privacy risks to their users~\cite{tiboni2022sensors, netukova2022lower}. The relatively accessible price point of recreational  devices in particular, a difference of thousands of dollars versus tens or even hundreds of thousands of dollars, significantly broadens the consumer base and emphasizes the need for proactive assessment of privacy risks.  


\section{Future Directions}
\label{sec: future direction}
Idiobionics presents an opportunity to explore new research ventures that strengthen the privacy posture of present and future robotic prostheses.
An open avenue for investigation is determining whether the very features that enable autonomous and adaptive capabilities simultaneously introduce privacy vulnerabilities. 
 This suggests a set of related questions for future research:

\begin{itemize}
    \item Are there other embedded sensors, actuators, computing components, or algorithmic components in bionic limbs that could be exploited to expose sensitive user data?
    \item Could the machine learning models used in many commercial prostheses, or their supporting software ecosystem, inadvertently leak private information about users? 
    \item  How can the new knowledge gained from identifying privacy threat vectors inform the creation of better hardware and software paradigms that are capable of providing robust privacy guarantees within autonomous and adaptive bionic devices?
    \item To what extent can existing Privacy-Preserving Machine Learning (PPML) techniques be applied or translated effectively to this domain?
    \item Is there a need to develop new, tailored PPML methodologies that can successfully balance the imperative of high accuracy which is critical for bionic device function with the provision of adequate practical privacy guarantees?
\end{itemize}

Exploring these research questions allows us to begin addressing both current and emerging privacy issues within autonomous wearable robotic systems. Although technical evaluations are crucial for identifying vulnerabilities and proactively mitigating risks, addressing privacy issues in intelligent bionic technologies requires more than just technical insight. Without insight from  actual end-users, we risk creating solutions that lack practical utility. There are further opportunities within idiobionics research to gain insight from users on their perception of privacy as it relates to their integrated bionic limbs, such as investigating: 
\begin{itemize}
    \item How do users of adaptive, machine learning-enabled devices attached to their bodies understand privacy?
    \item How can we provide privacy guarantees within these devices that satisfy user expectations?
    \item When and how do users balance the accuracy and functionality of their bionic limbs with their personal data privacy?
\end{itemize}

Prioritizing idiobionics research now establishes a foundation that will allow us to accelerate the development of intelligent robotic prostheses with confidence, such that users can fully reap the benefits of these technologies without fear of privacy violations.



\section{Conclusion}
In this paper we contribute idiobionics as a necessary and emergent research direction at the intersection of privacy and  bionic limbs. Our motivation for the advocacy of privacy in bionic limbs via this new line of inquiry is to guide the design of bionic limbs towards systems that users can trust, confidently rely on, and fully benefit from. As an example of idiobionics research, we demonstrated the feasibility of a malicious attacker inferring activities that might be performed by users with transradial bionic limbs.
The ease of deploying the adversarial methods in our empirical evaluation, and the associated privacy implications for users of increasingly complex bionic system, underscore the necessity for this new line of inquiry. As such, we also provided a set of concrete research questions as starting points for both technical and sociotechnical innovation. Ultimately, we believe idiobionics will allow us to unlock the full potential of the next generation of autonomous bionic limbs and intelligent IoB devices.

\bibliographystyle{IEEEtranS}
\bibliography{references}

\end{document}